\documentclass[journal]{IEEEtran}
\pdfoutput=1
\usepackage{cite}
\usepackage{amsmath,amssymb,amsfonts}
\usepackage{algorithmic}
\usepackage{graphicx}
\usepackage{textcomp}
\usepackage{comment}
\usepackage{xcolor}
\usepackage{array}
\usepackage{amsmath,amssymb,amsfonts}
\usepackage{amsthm}
\usepackage{mathrsfs}
\usepackage{CJK}
\usepackage[ruled,linesnumbered]{algorithm2e}
\usepackage{amsmath}
\usepackage{url}
\usepackage{color}
\usepackage{pifont}
\usepackage{balance}

\newtheorem{assumption}{Assumption}
\newtheorem{lemma}{Lemma}
\newtheorem{theorem}{Theorem}
\newtheorem{definition}{Definition}

\newcommand{\tabincell}[2]{\begin{tabular}{@{}#1@{}}#2\end{tabular}}

\allowdisplaybreaks[4]

\ifCLASSINFOpdf
\else
\fi
\begin{document}
\title{A Fast  Blockchain-based Federated Learning Framework with Compressed Communications}

\author{\IEEEauthorblockN{Laizhong Cui, Senior Member, IEEE,
Xiaoxin Su, Student Member, IEEE,
Yipeng Zhou\IEEEauthorrefmark{1}, Member, IEEE}

\IEEEauthorblockA{
\thanks{
\newline Manuscript received February 7, 2022; revised June 11, 2022; accepted
June 30, 2022. This work has been partially supported by National Key R\&D Program of China under Grant No.2018YFB1800302, National Natural Science Foundation of China under Grant No.61772345, Shenzhen Science and Technology Program under Grant No. RCYX20200714114645048, No. JCYJ20190808142207420 and No. GJHZ20190822095416463.
\newline L. Cui, X. Su, are with the College of Computer Science and Software Engineering, Shenzhen University, Shenzhen 518060, China (email: cuilz@szu.edu.cn; suxiaoxin2016@163.com).
\newline Y. Zhou is with the School of Computing, FSE, Macquarie University, Macquarie Park, 2113, NSW, Australia (email: yipeng.zhou@mq.edu.au).
\newline Y. Zhou is the corresponding author.
}
}
} 
\markboth{}{Shell \MakeLowercase{\textit{et al.}}: Bare Demo of IEEEtran.cls for IEEE Journals}
\maketitle

\begin{abstract}
Recently, blockchain-based federated learning (BFL) has attracted intensive research attention due to that the training process is auditable and the  architecture is serverless avoiding the single point failure of the parameter server in vanilla federated learning (VFL). Nevertheless, BFL tremendously escalates the communication traffic volume because all local model updates (i.e., changes of model parameters) obtained by BFL clients will be transmitted to all miners for verification and to all clients for aggregation. In contrast, the parameter server and clients in VFL only retain aggregated model updates.  Consequently, the huge communication traffic in BFL will inevitably impair the training efficiency and hinder the deployment of BFL in reality. To improve the practicality of BFL, we are among the first to propose a fast blockchain-based communication-efficient federated learning framework by compressing communications in BFL, called BCFL. Meanwhile, we derive the convergence rate of BCFL with non-convex loss.  To maximize the final model accuracy, we further formulate the problem to minimize the training loss of the convergence rate subject to a limited training time with respect to the compression rate and the block generation rate, which is a  bi-convex optimization problem and can be efficiently solved.
To the end, to demonstrate the efficiency of BCFL, we carry out extensive experiments with standard   CIFAR-10 and FEMNIST datasets. Our experimental results not only verify the correctness of our analysis, but also manifest that BCFL can remarkably reduce the communication traffic by 95-98\% or shorten the training time by 90-95\% compared with BFL.   

\end{abstract}

\begin{IEEEkeywords}
Federated Learning, Blockchain, Compression, Convergence
\end{IEEEkeywords}

\IEEEpeerreviewmaketitle

\section{Introduction}

To mitigate the rising concern on data privacy leakage, the federated learning (FL) paradigm emerges which can conduct model training without touching raw data samples residing on decentralized clients \cite{mcmahan2017communication}. In vanilla FL (VFL), a parameter server (PS) assists all clients in model aggregation. In each round of global iteration, the PS will distribute the latest model to selected clients before clients update the model by conducting a few number of local iterations with their local datasets. Then, updated model updates (\emph{i.e.}, changes of model parameters) are collected and aggregated by the PS. Multiple global iterations are  conducted by involving different clients until the model converges \cite{li2019convergence, yang2021achieving}.

Although, VFL can prevent the disclosure of  original data samples, it confronts the following two challenges in that VFL is highly dependent on the honesty and reliability of clients and the PS. First, the PS needs to communicate with multiple clients simultaneously, which probably chokes the PS such that the training process halts \cite{pokhrel2020federated}. Second, if the PS or clients maliciously tamper model updates or injecting forged samples, the model accuracy can be significantly compromised \cite{bagdasaryan2020backdoor}.

To overcome the deficiencies of VFL,  blockchain-based federated learning (BFL) was devised by \cite{kim2019blockchained}. On one hand, BFL is more robust which is invulnerable to the failure of a single point \cite{otoum2020blockchain}. On the other hand, all intermediate model updates generated by clients are auditable and traceable which can effectively prohibit adversarial behaviours \cite{bao2019flchain}.  
For instance, Kim \emph{et. al.} designed BlockFL \cite{kim2019blockchained} with a decentralized FL training mode combined with blockchain. 
Pokhrel \emph{et. al.}  \cite{pokhrel2020federated} studied how to use channel dynamic to minimize blockchain-based FL training delays in autonomous vehicle scenarios. Feng \emph{et. al.} proposed BAFL \cite{9399813} to  ensure the security and efficiency of federated learning through blockchain and asynchronous training.

Despite these advantages, BFL suffers from the long communication latency and the low training efficiency because of its huge communication traffic, which can be attributed to: 1) In BFL, all intermediate model updates will be broadcasted to all miners to maintain the blockchain; 2) Each client needs to download all intermediate model updates to locally conduct model aggregation. In contrast, only aggregated model updates are maintained by the PS and downloaded by clients in VFL, implying that the communication load of VFL is much lighter than that of BFL. 
It was reported in \cite{nguyen2021federated} that the communication traffic of BFL is too heavy making BFL impracticable in real systems. Several existing works have attempted to prohibit communication overhead in BFL. Xuan \emph{et. al.} \cite{xuan2021dam} proposed  a communication cost optimization method by proposing the verification mechanism to exclude useless or malicious nodes to reduce communication cost in BFL. In \cite{wilhelmi2021blockchain}, Wilhelmi \emph{et. al.} proposed an analytical model based on batch service queue theory to implement asynchronous training in BFL in order to reduce the block size and optimize communication latency. Yet,  
 compressing model updates to alleviate the communication load of BFL has not been explored by existing works.



In light of inefficient communications in BFL, we  are among the first to propose a fast blockchain-based communication-efficient federated learning framework with compressed communications, called BCFL. Intuitively speaking, the communication load in BFL is proportional to the size of intermediate model updates contributed by clients. Hence, the communication efficiency of BFL can be substantially improved if we can effectively shrink the population of intermediate model updates by discarding ones inessential for model training. BCFL leverages the $Top_k$ algorithm \cite{stich2018sparsified} (one of the most effective compression algorithms in federated learning) to compress local model updates on each client in order to alleviate the communication load of BCFL. Specifically, each BCFL client only injects $k$ most significant model updates into BCFL, where $k$ can be much smaller than the model dimension. The significance of each model update is determined by its absolute value. A model update close to $0$ is regarded as an inessential one for model training which only slightly affects the aggregated model \cite{aji2017sparse}.
In addition, we prove the convergence of BCFL with non-IID sample distribution and non-convex loss. To maximize the final model accuracy,  we formulate the problem to  minimize the convergence rate subject to a fixed training time span with respect to the compression rate and the block generation rate of BCFL. We prove that this  is a bi-convex optimization problem, which can be solved efficiently. In the end, the extraordinary performance of BCFL is demonstrated by experiments conducted with  the standard CIFAR-10 and FEMNIST datasets. 

In a word, the contributions of our work are unfolded as follows.
\begin{itemize}
\item To alleviate the communication load in BFL, we are among the first to propose the BCFL framework, which employs the {$Top_k$} algorithm to compress model updates. 
In addition,  We derive the convergence rate of BCFL under non-IID sample distribution and non-convex loss.
\item We explore how to optimally set the compression rate and the block generation rate in BFCL. Given a fixed training time span, we formulate the problem to minimize the loss function with respect to the compression rate and the block generation rate as a bi-convex optimization problem, which can be easily solved. 
\item We conduct comprehensive experiments to evaluate the performance of BCFL under both IID and non-IID  data distributions with CIFAR-10 and FEMNIST datasets. 
The experimental results manifest that BCFL can shrink the communication traffic by 95-98\%, or shorten the training time by 90-95\% compared with BFL without compression. 
\end{itemize}

The rest of the paper is organized as follows. Relevant  works are discussed in Sec.~\ref{RelatedWork}. Sec.~\ref{Preliminary} introduces the preliminary knowledge of federated learning. Then BCFL framework is elaborated in Sec.~\ref{BCFLFramework}. The convergence rate of BCFL  and the optimization of model accuracy are analyzed in Sec.~\ref{ConvergenceAnalysis}. Experimental results used to evaluate BCFL  performance are presented  in Sec.~\ref{Performance}. Ultimately, we conclude our work  in Sec~\ref{Conclusion}.

\section{Related Work} \label{RelatedWork}

In this section, we discuss related works from three aspects: FL, combination of blockchain and FL, and model compression in FL. 

\subsection{Federated Learning}
The federated average algorithm (FedAvg) \cite{mcmahan2017communication} is the most fundamental FL algorithm that can realize privacy preserved distributed model training. To date, the convergence of FedAvg algorithm has been analyzed in \cite{li2019convergence} with strongly convex loss and \cite{yang2021achieving}  with  non-convex loss,  respectively. Later on, variants of FedAvg are devised to accommodate constrained resources on clients.  Based on the convergence property of FL, Wang \emph{et al.} dynamically adjusted local training epochs on clients to minimize the loss function under restricted computation and communication resources \cite{wang2019adaptive}. FEDL \cite{dinh2020federated} modeled the resource allocation problem, and decomposed the obtained non-convex problem to optimize the convergence of the model. Hu \emph{et al.} \cite{hu2019decentralized} designed a model segmented gossip approach to achieve decentralized federated learning to fully utilize the bandwidth among nodes.
Luo \emph{et al.} \cite{luo2020cost} designed a multi-variable optimization problem by taking into account the learning time and energy consumption in FL model training, in order to minimize the total loss. 

\subsection{Combination of FL and Blockchain}

Blockchain is decentralized, non-tamperable, and highly transparent, which can naturally suit federated learning. A bunch of existing works have explored to combine blockchain with FL towards building a more robust and reliable model training architecture.

Biscotti \cite{9292450} was proposed as a fully decentralized peer-to-peer multi-party learning framework that ensures data security and privacy with the assistance  of blockchain. 
In addition, the  end-to-end delay of BFL is analyzed to derive the optimal block generation rate. In \cite{9127823},  a hierarchical blockchain framework with FL was proposed for knowledge sharing in large-scale vehicular networks.
In \cite{lu2020low}, Lu \emph{et. al.} proposed to  reduce the communication latency of BFL by integrating digital twin into wireless network to reduce unreliable communication between users and servers.
In \cite{zhao2021blockchain}, the authors designed a blockchain-based federated learning platform, which implements gradient aggregation by using smart contracts to reduce the risk of data privacy leakage.
ChainsFL \cite{yuan2021chainsfl} was proposed as  a two-layer blockchain-driven FL framework to deal with the problems of high resource consumption, limited throughput, and high communication complexity in BFL.
In spite of these efforts targeting to improve BFL efficiency, the communication load of BFL is still too heavy in practice.


\subsection{Communication Compression in FL}

Considering limited bandwidth resources and increasingly complex machine learning models in FL, tremendous efforts have been dedicated to improve the communication efficiency of FL by compressing communications.
Existing compression algorithms can be generally divided into two types: sparsification and quantization.

Sparsification compression algorithms filter  model updates to be transmitted and only retain important updates for transmission.
The DGC algorithm proposed in \cite{lin2017deep} discarded 99.9\% of insignificant model updates for uploading on distributed machine learning workers, and different methods were designed to maintain the performance of the model. 
ClusterGrad \cite{cui2020clustergrad} was based on clustering to adaptively filter and quantify important gradients according to the distribution of model updates.
Li \emph{et al.} \cite{li2020talk} took into account the heterogeneous resources on devices in FL. An optimization algorithm was proposed accordingly to adjust the compression rate of each device and local computation in order to minimize the total energy consumption.
DC2 \cite{abdelmoniem2021dc2} was a heuristically designed adaptive compression scheme that can adjust the amount of data to be transmitted according to the variation of network delay. Sattler \emph{et al.} proposed the STC algorithm \cite{sattler2019robust} to sparse the transmitted data and quantize the sparsed updates into two discrete values during upload and download processes, which thereby greatly reduces communication time. 
The quantization based compression algorithms quantize model updates to be transmitted to a number of discrete values with compromised parameter precision. 
In \cite{alistarh2017qsgd}, the proposed QSGD algorithm generated a random number for each transmitted update, and used it to map the update to a centroid in an unbiased manner.
PQ algorithm proposed in \cite{suresh2017distributed} divided the difference between maximum and minimum values of model updates into equidistant intervals, and randomly quantized the updates in each interval into its upper or lower bound.
In \cite{wen2017terngrad}, the TernGrad algorithm was designed to ternary model updates to improve the convergence of the algorithm by layer-wise ternarizing and gradient clipping.
Cui \emph{et. al.} proposed MUCSC \cite{cui2021slashing} to quantify model updates by analyzing the effect of compression error on model convergence, which can correspondingly select  centroids to minimize the error.

Although compressing communications have been widely discussed in FL, there is little effort to apply model compression techniques in BFL and this gap will be bridged by our work.

\section{Preliminary} \label{Preliminary}

To pave the discussion of our BCFL framework, we first briefly explain the FL training process with the most fundamental FedAvg algorithm. In FL, data samples are distributed across multiple clients. Without loss of generality, we assume that there are a total of $N$ clients and the dataset on client $i$ is $\mathcal{D}_i$. The local loss function of client $i$ is defined as
\begin{equation}
F_i(\mathbf{w}, \mathcal{D}_i)=\frac{1}{|\mathcal{D}_i|}\sum_{\xi \in \mathcal{D}_i}f(\mathbf{w}, \xi),
\end{equation}
where $\mathbf{w}$ represents  model parameters, $|\mathcal{D}_i|$ is the size of local dataset, $\xi$ is a particular data sample and $f()$ is the loss function of a particular machine learning task. The goal of FL is to train a model that minimizes the global loss function, \emph{i.e.,}
\begin{equation}
\label{EQ:GlobalLoss}
\min_{\mathbf{w}} F(\mathbf{w})= \min_{\mathbf{w}} \sum_{i=1}^N p_iF_i(\mathbf{w}, \mathcal{D}_i),
\end{equation}
where $p_i$ is the weight of client $i$, and it is usually defined as $p_i=\frac{|\mathcal{D}_i|}{\sum_{i^{'}=1}^N|\mathcal{D}_{i^{'}}|}$. 

The objective of FedAvg is to train the model to minimize the loss defined in Eq.~\eqref{EQ:GlobalLoss}.  FedAvg consists of multiple global iterations. In global iteration $t$, the parameter server (PS) randomly selects $K$ clients as $\mathcal{K}_t$ to participate the $t$-th global iteration.
The selected client $i$ will download the latest model $\mathbf{w}^i_{t}=\mathbf{w}_{t}$ from the PS, and then performs $E$ local iterations (a.k.a epochs). The gradient descent with a mini-batch derived in each local iteration is as follow
\begin{equation}
\mathbf{w}^i_{t+1} = \mathbf{w}^i_{t} - \eta \nabla F_i(\mathbf{w}^i_{t}, \mathcal{B}^i_t),
\end{equation}
where $\mathcal{B}^i_t$ is a mini-batch of samples with size $b$ selected from $\mathcal{D}_i$. Let $\mathbf{g}_{t+E}^i$ denote model updates of client $i$ by conducting $E$ local epochs in the $t$-th global iteration, which can be exactly defined as $\mathbf{g}_{t+E}^i = \mathbf{w}_{t+E}^i - \mathbf{w}_{t}^i$.  After conducting $E$ local iterations,  selected clients upload model updates  to the PS for the following global aggregation
\begin{equation}
\label{EQ:GlobalAgg}
\mathbf{w}_{t+E} = \mathbf{w}_{t} + \sum_{i\in\mathcal{K}_t}p_i \mathbf{g}_{t+E}^i.
\end{equation}
With globally updated model parameters obtained in Eq.~\eqref{EQ:GlobalAgg}, the PS can embark a new round of global iteration by involving different participating clients. 

Note that model updates are formally defined as changes of model parameters  after conducting $E$ local epochs on clients. In other words, model updates from client $i$ are $\mathbf{g}_{t+E}^i=\sum_{j=t}^{t+E-1}\eta\nabla F_i(\mathbf{w}^i_j, \mathcal{B}^i_j)$. Given that clients and the PS merely exchange model updates during model training, our work focuses on compressing model updates. 

For simplicity, let $t$ denote the index of all iterations and there are total $T$ iterations.  The global model will be updated only if $t\in\mathcal{I}=\{E, 2E, 3E, \dots\}$. Note that the initial model parameters can be randomly generated by the PS when $t=0$.

\section{BCFL Framework}\label{BCFLFramework}

In this section, we elaborate the training process of the BCFL framework to illustrate the relation between the training time cost of BCFL and the amount of communication traffic. Then, we present the  BCFL algorithm by leveraging the $Top_k$ compression algorithm. 

\subsection{BCFL Training Process}


\begin{figure}[h]
\centering
\includegraphics[width=\linewidth]{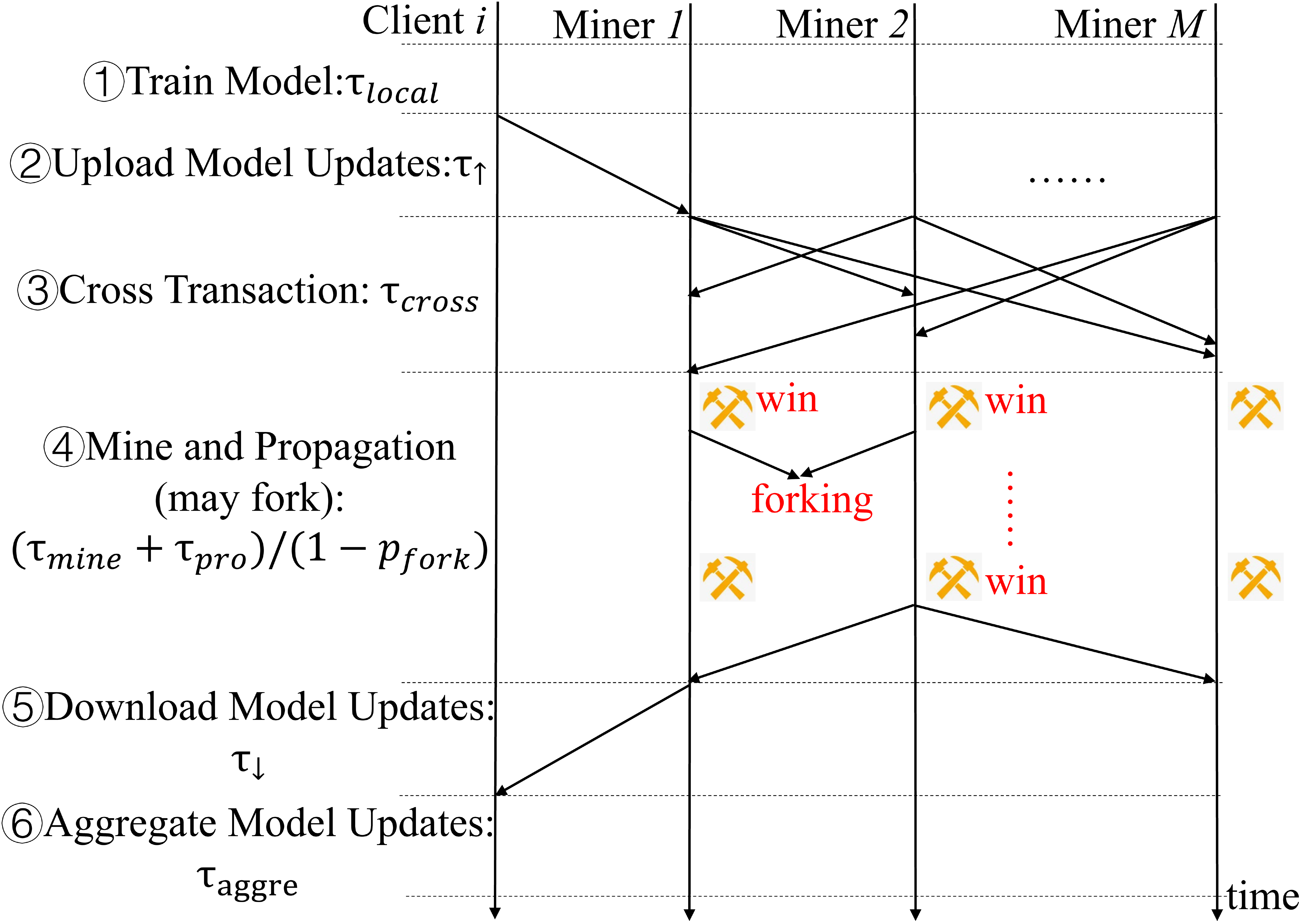}
\caption{ {BCFL Framework Process} }
\label{Framework_Process}
\end{figure}



Similar to previous works,  we consider that there are $N$ clients and $M$ miners in the blockchain-based learning system. Let  $\mathcal{N}$ and $\mathcal{M}$ denote the set of clients and miners, respectively. We suppose that each client is connected to a particular miner in each global iteration to upload and download model updates.

Note that BCFL is a generic framework since different compression algorithms and consensus mechanisms can be adopted. 
However, to analyze the convergence rate of BCFL, we specify the $Top_k$ algorithm \cite{han2020adaptive} for model update compression and PoW (proof of work) as the consensus mechanism hereafter.   PoW is the most fundamental  consensus mechanism \cite{nakamoto2008bitcoin} used in blockchain, and also widely used in BFL \cite{9664296, 9399813}. It is not difficult to adopt a different  consensus mechanism  in BCFL by modifying the mining model accordingly.   

In BCFL, due to the involvement of miners and the distributed aggregation operated on each client, the training process of BCFL is much more complicated than that of VFL. There are six steps in a round of global iteration. 
Next, we will introduce each step in details.

\begin{enumerate}
\item{\bf Local Model Training}  Each client uses batch gradient descent to update local model parameters according to $\mathbf{w}^i_{t+\frac{1}{2}}=\mathbf{w}^i_{t}-\eta\nabla F_i(\mathbf{w}^i_{t}, \mathcal{B}^i_t) $ where $\mathcal{B}^i_t$ is a batch of samples with size $b$ selected from  $\mathcal{D}_i$.  
If the iteration index $t+1 \notin \mathcal{I}$, it is not a global synchronization iteration, and thus  $\mathbf{w}^i_{t+1}= \mathbf{w}^i_{t+\frac{1}{2}}$.
Otherwise, if $t+1\in \mathcal{I}$, it means that  the client has completed $E$ local iterations and  needs to upload local model updates to its miner. 
The model updates defined as $\mathbf{g}_{t}^i = \mathbf{w}_{t+\frac{1}{2}}^i-\mathbf{w}_{t+1-E}- \mathbf{m}^i_{t+1-E}$ will be compressed by the $Top_k$ algorithm as below. 
\begin{equation}
Top_k(\mathbf{g}_{t}^i[l])=\left\{
\begin{aligned}
&\mathbf{g}_{t}^i[l], \quad \textit{$|\mathbf{g}_{t}^i[l]| > \phi$},\\
&0, \quad\quad\quad\quad \textit{otherwise}, \\
\end{aligned}
\right.
\end{equation}
where $\phi$ is the threshold determined by the value of the $(k+1)$-th largest absolute value of items in $\mathbf{g}_{t}^i$ and $\mathbf{g}_{t}^i[l]$ is the $l$-th item of $\mathbf{g}_{t}^i$.
Here $\mathbf{m}^i_{t+1-E} $ is the compensation error which is used to locally adjust model parameters based on the compression error. The compensation error is defined as  $\mathbf{m}^i_{t+1}=Top_k(\mathbf{g}^i_{t}) + \mathbf{w}^i_{t+\frac{1}{2}} - \mathbf{m}^i_{t+1-E} - \mathbf{w}_{t+1-E}$. 
Intuitively speaking, the $Top_k$ algorithm only uploads $k$ most significant items. 
Let $\tau_{local}$ denote the time spent on this step in BCFL.

\item {\bf Model Upload} 
The upload time of model updates from a particular client to its miner is determined by the traffic volume and the upload speed. Let $d$ denote the  dimension of the model. Suppose that the traffic volume of each model update is $s$ bytes. The traffic volume of $k$ updates is  $k(s+\frac{\lceil \log_2 d \rceil}{8})$ bytes, where $k$ is the number of selected updates and $\frac{\lceil \log_2 d \rceil}{8}$ is the number of bytes to represent indices of these updates. In contrast, the traffic volume will be  $ds$ in VFL with  no compression. Let $u_{\uparrow,i}$ bytes/s denote the upload speed of client $i$, then the elapsed time of this step is $\tau_{\uparrow,i} = \frac{k(s+\frac{\lceil \log_2 d \rceil}{8})}{u_{\uparrow,i}}$.  
Note that the remaining $d-k$ insignificant model updates will not be uploaded by client $i$, which will be absorbed by the compensation error $\mathbf{m}_{t+1}^i$.

\item {\bf Cross Transaction}  In this step, all model updates uploaded by clients will be stored in candidate blocks. Thereafter,  each miner needs to distribute transactions (\emph{i.e.}, model updates) received from its clients to all other miners.   According to \cite{kim2019blockchained}, the bottleneck for distributing transactions lies in the download capacity of miners. Let $u_j$ denote the download speed of miner $j$. Let $\mathcal{N}_j$ and $N_j$ denote the set of and the number of clients  clients connecting miner $j$, respectively. It takes  $\tau_{cross,j}=\frac{(N-N_j)k(s+\frac{\lceil \log_2 d \rceil}{8})}{u_j}$ for miner $j$ to receive transactions from all other miners, where $N-N_j$ represents the number of clients not connecting to miner $j$.  

\item {\bf Mine and Propagation} 
In this step, each miner attempts to generate a block to store all model updates from clients into the blockchain. However, it is not trivial to derive the elapsed time of this step because: 1) The elapsed time is a random variable such that we can only analyze its expected value; 2) We need to consider both the time to generate and propagate a block, and the fork probability. Once a fork event occurs, this step has to start over to consume excessive time.  
In practice,  miners will not receive transactions in a synchronized manner.  However, according to previous works \cite{kim2019blockchained, deng2021dynamic}, this difference is negligible. To simplify analysis, we ignore this difference in our work.

We first consider the expected elapsed time for a winning miner to generate a block (denoted by $\tau_{mine}$) and propagate the block to all miners (denoted by $\tau_{pro}$) without considering the occurrence of forks. The process for the winning miner to generate a block is determined by consensus mechanisms. 
In this work, we adopt the PoW consensus mechanism for analysis. With PoW mechanism, each miner enumerates different \emph{nonce} and hashes the block header until the hash value of one miner is less than a given target value. Then this miner obtains the right to connect its candidate block to the blockchain and broadcast it to other miners. Other miners stop PoW processes once they receive the transmitted block and connect it to their own blockchains. Note that  different consensus mechanisms can be adopted for miners to compete for the right of bookkeeping. We only need to alter $\tau_{mine}$ slightly if a different consensus mechanism is adopted.   

According to \cite{nakamoto2008bitcoin}, if PoW is adopted as the consensus mechanism, the target value can be tuned flexibly by  adjusting the difficulty to obtain the hash result, and the process to generate a block can be modeled by a random variable obeying exponential distribution. Let random variable $x_{mine}$ denote the elapsed time of this process. Let $\lambda$ denote the parameter of the exponential distribution, then the expected mining time  is $\tau_{mine} = E[x_{mine}]=\frac{1}{\lambda}$ where $\lambda$ is a tunable parameter. How to optimally set $\lambda$ will be discussed later. 

Let $j_w$ denote the first miner to get a block. The miner $j_w$ needs to send its block to all other miners. The block size is $\Omega = Nk(s+\frac{\lceil \log_2 d \rceil}{8})$ bytes containing model updates of all clients. Recall that the miner set is denoted by $\mathcal{M}$. Then, the elapsed time for mining and propagation is 
$\tau_{mine}+\tau_{pro} = \frac{1}{\lambda} + \max_{j \in \mathcal{M}/j_w} \frac{\Omega}{u_j}$. Here $\mathcal{M}/j_w$ denotes the set of all miners except miner $j_w$.

Nonetheless, forks occur if any other  miners also generate blocks before $j_w$ can propagate its block to these miners. Once forks occur, this step starts over. By modeling $x_{mine}$ with exponential distribution, the fork probability can be estimated accurately. Let $x_{mine, j}$ denote the time for miner $j$ to generate a block with the same distribution as $x_{mine}$.  
To avoid forking, the  block generated by miner $j_w$  should arrive at miner $j$ before miner $j$ obtains a block. 
In other words, it is required that $x_{mine,j}-x_{mine,j_w}> \frac{\Omega}{u_j} $. Therefore, the forking probability is  $p_{fork}=1-\prod_{j\in\mathcal{M}/j_w}Pr\left(x_{mine,j}-x_{mine,j_w} > \frac{\Omega}{u_j} \right) $. Here  we assume that all miners start competing simultaneously. In avoid forking, we must have
\begin{align*}
&Pr\Big(x_{mine,j}-x_{mine,j_w} > \frac{\Omega}{u_j} \Big)\notag\\
&=Pr\Big(x_{mine,j} > x_{mine,j_w} + \frac{\Omega}{u_j} | x_{mine,j} > x_{mine,j_w}\Big)\notag\\
&=Pr\Big(x_{mine,j} > \frac{\Omega}{u_j}\Big) =   \exp(-\lambda \frac{\Omega}{u_j}).
\end{align*}
Here, the derivation is based on the memorylessness of the exponential distribution of $x_{mine, j}$.
Consequently, the probability of forking is $p_{fork}=1-\exp(-\lambda \sum_{i\in\mathcal{M}/j_w}\frac{\Omega}{u_j})$.

By wrapping up, the expected elapsed time of this step is 
\begin{equation}
\begin{aligned}
&(\tau_{mine}+\tau_{pro})/(1-p_{fork})\\
&=\left(\frac{1}{\lambda}+\max_{j \in \mathcal{M}/j_w} \frac{\Omega}{u_j}\right) \exp\left(\lambda \sum_{j\in\mathcal{M}/j_w}\frac{\Omega}{u_j}\right).
\end{aligned}
\end{equation}

\item {\bf  Model Download} When a block is successfully generated by the winning miner, all clients need to download the latest block containing model updates from all clients to update the trained model. 
Let $u_{\downarrow,i}$ denote the download speed of client $i$. 
The download time consumed by client $i$ is $\tau_{\downarrow,i}=\frac{\Omega}{u_{\downarrow,i}}$.

\item {\bf Model Aggregation} Once clients obtain the latest block, they can aggregate model updates to deduce updated model parameters. Let $\tau_{aggre}$  denote the time consumed by the simple aggregation operation, which can be easily estimated  based on the computation capacity of clients. 
\end{enumerate}

\noindent\textbf{Discussion:} Based on the analysis of the training process of BCFL, we can conclude that the time cost of a round of global iteration is the cumulative time cost of aforementioned six steps, which is 
\begin{equation}
\label{EQ:TimeCost}
\begin{aligned}
h(k,\lambda)&=\tau_{local}+\tau_{aggre}+\max_{i \in \mathcal{N}} \tau_{\uparrow,i} + \max_{j \in \mathcal{M}}\tau_{cross,j}\\
&\quad+(\tau_{mine}+\tau_{pro})/(1-p_{fork})+\max_{i \in \mathcal{N}} \tau_{\downarrow,i},
\end{aligned}
\end{equation}
where $k$ is the parameter of the $Top_k$ compression algorithm. The explanation of each symbol for the time cost analysis is listed in Table~\ref{NotationListOfTau} in Appendix. 

\noindent {\bf Remark I:} We can suppose that the computation and communication capacity of each client and miner can be measured in advance. Given the volume of communication traffic such as $\Omega$, the time cost $h(k,\lambda)$ can be estimated accurately prior to the commencement of BCFL. 

\noindent {\bf Remark II:} The time cost $h(k,\lambda)$  is heavily affected by the traffic volume $\Omega$. If $k$ is smaller, it implies that the time cost can be reduced effectively owing to smaller $\Omega$. Consequently, BCFL can complete a round of global iteration with a faster speed. 

\noindent {\bf Remark III:} However, it is unreasonable to merely reduce $k$ without considering its influence on the convergence of the trained model. To optimally set the compression rate in BCFL, we conduct convergence analysis in the next section to synthetically consider the time cost and the convergence rate so as to maximize the final model accuracy within a fixed span of training time. 



\subsection{BCFL Algorithm}
To have a global picture of the BCFL framework and facilitate subsequent convergence analysis, we present the BCFL algorithm in Alg.~\ref{tranProcess}. To make our presentation uncluttered, the step number has been marked as comments in the algorithm. 
Different from VFL, there is no PS in  BCFL. Instead, a number of miners engage in the training process. A client can participate FL  as long as the client connects with any  miner.


\begin{algorithm}[t]
\label{tranProcess}
\caption{BCFL Framework Training Process}
\LinesNumbered
\KwIn{Model $\mathbf{w}_0, \mathbf{m}^i_0=\mathbf{0}, \forall i \in [N]$; learning rate $\eta$}
\KwOut{Final model $\mathbf{w}_T$}
Each client is randomly connected to a miner.\\
\For{$t=0$ to $T-1$}{
\textbf{On Clients:} \\
\For{$i=1$ to $N$ \textbf{parallel}}{
\tcp{Lines 5 to 8 correspond to Step 1)}
$\mathbf{w}^i_{t+\frac{1}{2}}=\mathbf{w}^i_{t}-\eta\nabla F_i(\mathbf{w}^i_{t}, \mathcal{B}^i_t) $.\\
\If{$t+1 \notin \mathcal{I}$}{
$\mathbf{w}^i_{t+1}=\mathbf{w}^i_{t+\frac{1}{2}}$.\\
}
\Else{
\tcp{Correspond to Step 2)}
Send $\mathbf{g}^i_{t}=Top_k(\mathbf{w}^i_{t+\frac{1}{2}} - \mathbf{m}^i_{t+1-E}-\mathbf{w}_{t+1-E})$ to miner. \\
$\mathbf{m}^i_{t+1}=\mathbf{g}^i_{t} + \mathbf{w}^i_{t+\frac{1}{2}} - \mathbf{m}^i_{t+1-E} - \mathbf{w}_{t+1-E}$.\\
\tcp{Correspond to Step 5), 6)}
Download $\mathbf{g}^i_{t}, \forall i$ from miner and set $\mathbf{w}^i_{t+1}=\mathbf{w}_{t+1}=\mathbf{w}_{t+1-E}+\frac{1}{N}\sum_{i=1}^N\mathbf{g}^i_{t}$.
}
}
\textbf{On Miners:} \\
\For{$j=1$ to $M$ \textbf{parallel}}{
\If{$t+1 \in \mathcal{I}$}{
Receive $\mathbf{g}^i_{t}$ if $i \in \mathcal{N}_j$.\\
\tcp{Correspond to Step 3)}
Cross received $\mathbf{g}^i_{t}$ to miners $j^{'}$ if $j^{'} \neq j$. \\
\tcp{Lines 20 to 31 correspond to Step 4)}
\While{True}{
Execute consensus mechanism\\
\If{receive generated block}{
	Send $ACK$.\\
}
\ElseIf{generate block}{
	Send block to other miners.\\
}
\If{not forking}{
	break.\\
}
}
}
}
}
\Return{$\mathbf{w}_T$}
\end{algorithm}

\section{Convergence Analysis of BCFL}\label{ConvergenceAnalysis}

In this section, we  analyze the convergence of the BCFL algorithm and formulate the optimization problem to maximize the final model accuracy based on the derived convergence. 

\subsection{Assumptions and Definitions}
For the sake of  the popularity of neural network models \cite{lecun1998gradient, targ2016resnet}, we analyze the convergence of BCFL with non-convex loss. 
Similar to previous works \cite{gao2021convergence, yu2019parallel, dinh2020federated}, we make a few general assumptions.

\begin{assumption}
\label{Assump:ConSmoo}
The loss functions, \emph{i.e.}, $F_1, F_2,\dots, F_N$ are all $L$-smooth. In other words, given $\mathbf{v}$ and $\mathbf{w}$, we have $F_i(\mathbf{v}) \le F_i(\mathbf{w}) + (\mathbf{v}-\mathbf{w})^T\nabla F_i(\mathbf{w})+\frac{L}{2}||\mathbf{v}-\mathbf{w}||^2$.
\end{assumption}
Let $\xi^i_t$ denote the sample randomly and uniformly selected from client $i$.
\begin{assumption}
\label{Assump:LocalVar}
The variance of the stochastic gradients in each client is bounded, \emph{i.e.},
$\mathbb{E}[\|\nabla F_i(\mathbf{w}^i_t, \xi^i_t)-\nabla F_i(\mathbf{w}^i_t)\|^2] \le \sigma_i^2$ for $i=1, 2,\dots, N$ and $t = 0, 1,\dots, T-1$.
\end{assumption} 

\begin{assumption}
\label{Assump:BoundG}
The expected square norm of stochastic gradients is uniformly bounded, \emph{i.e.},  $\mathbb{E}[\|\nabla F_i(\mathbf{w}^i_t, \xi^i_t)\|^2] \le G^2$ for all $i = 1,2,\dots,N$ and $t = 0, 1,\dots, T-1$.
\end{assumption}

In most FL systems, the sample distribution on clients is non-IID, which can heavily impact the convergence of FL. In this work, we use the difference between local gradients and global gradients to quantify the non-IID degree of the sample distribution on  clients, which is also used in the previous work \cite{wang2019adaptive}.
\begin{definition} (Quantification of non-IID)
\label{QuantificationOfnon-IID}
We use $\Gamma_G^2 \ge \mathbb{E}[\|\nabla F_i(\mathbf{w}_t)-\nabla F(\mathbf{w}_t)\|^2], \forall i,t$ to quantize the degree of non-IID sample distribution on clients. 
\end{definition}
Note that the quantification becomes $0$ if the sample distribution is IID. 
It is common that the  distribution is non-IID for samples on decentralized clients in  FL. Consistent with prior works such as \cite{yang2021achieving, wang2019adaptive},  our analysis also shows that non-IID distribution can slow down the speed of model training in FL.
The adverse influence of non-IID can be mitigated by devising more advanced  client participation schemes \cite{9155494,zhao2018federated}. Note that this line of works is orthogonal to our work by considering that our focus is on optimizing the compression rate and block generate rate. 

The $Top_k$ algorithm used in the BCFL framework is a well-known biased compression algorithm \cite{stich2018sparsified, li2020talk}. The compression error is a major factor influencing the convergence, which is defined as follows. 
\begin{definition}
\label{CompressionOp}
$Top_k$ is a reasonable compression operation if there is a constant $\gamma$ such that the following compression property is satisfied,
$$\mathbb{E}[\|\mathbf{x}-Top_k(\mathbf{x})\|^2]\leq (1-\gamma)\|\mathbf{x}\|^2.$$
\end{definition}
The proof of this property has been derived in the previous work \cite{stich2018sparsified}. According to the derivation in \cite{stich2018sparsified},  $\gamma=\frac{k}{d}$ for the $Top_k$ algorithm where $d$ is the dimension of the trained model while $k$ is the number of reserved most significant model updates for communications.
It is worth noting that our framework is flexible in the sense that a 
different compression algorithm can be adopted here  based on system needs. 
In other words,  $\gamma $ in Definition~\ref{CompressionOp} should be updated accordingly if another compression algorithm is adopted. For example, $\gamma$ corresponding to STC is $\frac{\|Top_k(\mathbf{x})\|_1^2}{k\|\mathbf{x}\|^2}$ \cite{sattler2019robust}. 

\subsection{Convergence Rate}

We complete the proof of the convergence of BCFL by leveraging the  the convergence proof sketched in \cite{li2020talk, basu2020qsparse}. The difference lies in that \cite{li2020talk, basu2020qsparse} failed to consider the non-IID sample distribution  in FL. We extend this proof by taking the non-IID sample distribution into account. 

Let $\mathbf{z}_T$ denote the  uniformly and randomly selected sample of  historical  gradients  $\nabla\mathbf{w}^i_t$ for $\forall t, i$. In other words,   $Pr(\mathbf{z}_T=\nabla F(\mathbf{w}^i_t))=\frac{1}{NT}$ where $1\leq i\leq N$ and $1\leq t\leq T$. 
It turns out that the convergence rate of BCFL is: 
\begin{theorem}
\label{ConvergenceOfModel}
Let Assumptions~\ref{Assump:ConSmoo} to ~\ref{Assump:BoundG} hold. We set a fixed learning rate 
$\eta=\frac{{C}}{\sqrt{T}}$  and $\eta $ satisfies $\eta\le\frac{1}{16L}$, where ${C}$ is a constant. After conducting $T$ iterations with Algo.~\ref{tranProcess},
$	\mathbb{E}\|\mathbf{z}_T\|^2$ is bounded by:

\begin{equation}
\begin{aligned}
\mathbb{E}\|\mathbf{z}_T\|^2
&\le\left(\frac{\mathbb{E}[F(\mathbf{w}_{0})]- F^*}{C}+2C L\Gamma_G^2+\frac{C L}{bN^2}\sum_i\sigma_i^2\right)\frac{8}{\sqrt{T}}\\
&\qquad + \left(\frac{4}{\gamma^2}-3 \right)\frac{16C^2L^2G^2E^2}{T}.
\end{aligned}
\end{equation}
Here $F^*$ is the optimal value of global loss function.
\end{theorem}

Please refer to Appendix~\ref{ProofOfConvergence} for the detailed proof.
The key to the proof is to construct an auxiliary sequence of centrally trained models. We then analyze the model's updates in this sequence and its difference from the real trained model to derive the convergence. 
To ease the interpretation of our proof, the explanation of notations used for the convergence proof is listed in Table~\ref{NotationListOfTheorem} in Appendix. 

\noindent{\bf Remark I:}  
According to Theorem~\ref{ConvergenceOfModel}, the term $\frac{4}{\gamma^2}-3$ in the convergence rate is affected by the compression rate. Recall that the compression rate is $\gamma = \frac{k}{d}$. We can observe that  the convergence rate is lower in terms of the number of iterations if $\gamma$ is smaller (implying a higher compression rate). 

\noindent{\bf Remark II:}  On the other hand, our study in Eq.~\eqref{EQ:TimeCost} indicates that the time cost of each global iteration can be reduced by raising the compression rate $\gamma$. Therefore, it is not trivial to choose the optimal compression rate for BCFL. In view of that,  our next step is to  formulate the problem to optimize the final model accuracy with a fixed training time span by regarding the compression rate as a tuneable parameter.  


\subsection{Optimizing Model Accuracy}

In reality, the training time span for FL is limited. Let $Y$ denote the fixed training time span for FL. Then, our question is how to maximize the final model accuracy within the fixed training time span by tuning parameters $k$ and $\lambda$.

Recall that $k$ and $\lambda$  are two parameters affecting the time cost of one round of global iteration in $h(k, \lambda)$. To make our discussion concise, we let $\Lambda_A=\frac{8}{\sqrt{E}}\left(\frac{\mathbb{E}[F(\mathbf{w}_{0})]- F^*}{C}+2C L\Gamma_G^2+\frac{C L}{bN^2}\sum_i\sigma_i^2\right)$ and $\Lambda_B=16C^2L^2G^2E$ denote constant numbers not related with $k$ or $\lambda$ in the convergence rate.
Let $R = \frac{T}{E}$ denote the total number of conducted global iterations. The convergence rate in Theorem~\ref{ConvergenceOfModel} is simplified as
\begin{equation}
\label{EQ:SimpleConv}
\begin{aligned}
\mathbb{E}\|\mathbf{z}_T\|^2
&\le\frac{\Lambda_A}{\sqrt{R}} + \frac{\Lambda_B\left({\frac{4d^2}{k^2}-3} \right)}{R}.
\end{aligned}
\end{equation}





Given that the total training time span is $Y$, the number of conducted global iterations is $R=\frac{Y}{h(k,\lambda)}$. By substituting $R=\frac{Y}{h(k, \lambda)}$ into Eq.~\eqref{EQ:SimpleConv}, we can define our objective as:
\begin{equation}
\label{EQ:Obj}
\mathcal{J}(k, \lambda) =  \frac{\Lambda_A\sqrt{h(k,\lambda)}}{\sqrt{Y}} + \frac{\Lambda_B\left(\frac{4d^2}{k^2}-3 \right) h(k,\lambda)}{Y}.
\end{equation}
The problem to maximize the final model accuracy  (equivalent to minimizing the bound of $	\mathbb{E}\|\mathbf{z}_T\|^2$) is formally defined as:
\begin{eqnarray}
\label{EQ:optimV1}
\mathbb{P}1: &&\min_{k,\lambda} \mathcal{J}(k, \lambda)  \notag\\
&&\textit{s.t.} \quad 0<k\le d, \lambda>0 .
\end{eqnarray}
Considering that $d\gg k$ for compression algorithms in practice, the objective can be simplified as $\mathcal{J}(k, \lambda)  =\frac{\Lambda_A\sqrt{h(k,\lambda)}}{\sqrt{Y}} + \frac{\Lambda_B\frac{4d^2}{k^2}  h(k,\lambda)}{Y}$. Unfortunately, the objective function  $\mathcal{J}(k, \lambda) $ is not a convex or concave function. We resort to minimizing $\mathcal{J}^2 (k, \lambda)$. Our problem $\mathbb{P}1$ is converted to:  
\begin{eqnarray}
\label{EQ:optimV2}
\mathbb{P}2:  &&\min_{k,\lambda} \mathcal{J}^2 (k, \lambda) \notag\\
&&s.t. \quad 0<k\le d, \lambda>0\notag\\
&& \quad\quad\mathcal{J}^2 (k, \lambda) = \frac{\Lambda_A^2h(k,\lambda)}{Y} +\frac{16\Lambda_B^2d^4}{Y^2}\left(\frac{h(k,\lambda)}{k^2} \right)^2 \notag\\ &&\quad\quad+\frac{8\Lambda_A\Lambda_Bd^2}{Y^{\frac{3}{2}}} \left(\frac{h(k,\lambda)}{k^{\frac{4}{3}}} \right)^{\frac{3}{2}}.
\end{eqnarray}

We prove the following important property of $\mathbb{P}2$ by relaxing that $k$ is a positive real number. 
\begin{theorem}
\label{biConvex}
When $\lambda$ is fixed, $\mathcal{J}^2 (k, \lambda)$ is a convex function with respect to $k$. When $k$ is fixed, $\mathcal{J}^2 (k, \lambda)$ is a convex function with respect to $\lambda$. Given the linear constraints of $k$ and $\lambda$,  $\mathbb{P}2$ is a bi-convex optimization problem, which can be solved efficiently by Alternate Convex Search (ACS) \cite{gorski2007biconvex}.
\end{theorem}

Please refer to Appendix~\ref{ProofofBiConvex} for the detailed proof.
Therefore, $k$ and $\lambda$ in the BCFL algorithm can be determined by solving  $\mathbb{P}2$. In practice, we can set $k$  as the nearest integer of its  theoretically optimal value. 

\subsection{Practical Implementation}

From previous analysis, we can find that solving $\mathbb{P}2$ replies on the knowledge of network conditions such as the communication speed between miners so as to quantify the function $h(k, \lambda)$. Unfortunately, the network status is dynamic in practice, resulting in the change of $h(k, \lambda)$ and the solution of $\mathbb{P}2$. Thus, to implement BCFL, it is necessary to take the dynamic network conditions into account.


In fact, the network conditions can be proactively estimated and gauged through measurement \cite{lai1999measuring}. Due to the bi-convex property of $\mathbb{P}2$, we propose that the solution of $\mathbb{P}2$ can be updated in real time according to the latest network conditions. In other words, miners can periodically measure network conditions to update $h(k, \lambda)$, which is then substituted into  $\mathbb{P}2$ to update the compression rate and block generation rate. 


\section{Performance Evaluation} \label{Performance}
In this section, we  conduct experiments with CIFAR-10 and FEMNIST datasets to evaluate the performance of the BCFL framework.

\subsection{Experimental Settings}
\subsubsection{Dataset}
In our experiments, we employ CIFAR-10 and FEMNIST datasets for model training.
\begin{itemize}
\item CIFAR-10 includes 60,000 data samples, each of which is a 3*32*32 color image. This dataset contains ten labels, and each label consists of 6,000 samples. We randomly select 50,000 samples as the training set distributed on clients, and the remaining 10,000 samples will be used as the test set to evaluate the accuracy of the trained model.  Data samples are allocated to clients according to either  IID or non-IID distribution. For the former one, each client uniformly and randomly draws the same number of samples from the entire training set. For the latter one, each client randomly selects the same number of images from the training subset consisting of 5 random labels.  
\item FEMNIST \cite{caldas2018leaf} is a benchmark dataset for validating FL. Each sample is a handwritten 28*28 picture of digits and English characters, and there are 62 labels in total. The distribution is naturally non-IID since this dataset simulates the non-IID distribution of data samples  partitioned among different writers. In our experiments, each client owns more than 400 samples drawn from a single writer and 10\% of these samples will be allocated to the test set.
\end{itemize}
In summary, we have three sample distribution scenarios for experiments, which are CIFAR-10+IID, CIFAR-10+non-IID and FEMNIST. 
\subsubsection{Learning Model}
We train a convolutional neural network (CNN) model \cite{wang2020optimizing} to classify the CIFAR-10 dataset. The CNN model consists of 3 convolutional layers, each of which is 3*3 in size and the number of convolution kernels is 32, 64 and 64, respectively. Finally, two fully connected layers are used to output predictions for 10 labels. The model  has 122,570 parameters in total.

For the FEMNIST dataset, we refer to the model in \cite{caldas2018leaf} for training, which consists of two convolutional layers with max pooling layers and two fully connected layers. There are 111,902 parameters in total. The model  outputs predictions for 62 labels.

According to our analysis, we set a fixed learning rate as $\eta=0.05$ and $C=0.15$ for the training of both models. Because the number of data samples on each client is relatively small, we execute the full-batch gradient descent algorithm to conduct local iterations. Each client will perform $E=5$ local iterations in each global iteration. 

\subsubsection{Baseline Algorithm}
In our experiments, we compare the performance of BCFL with two kinds of baselines. The first one is the kind of BFL algorithms such as the one proposed in \cite{pokhrel2020federated} without compressing model updates for communications. We implement the one in \cite{kim2019blockchained} in our experiments, which only optimally sets $\lambda $ once the communication traffic is fixed.   
For the second kind, we randomly enumerate several different compression rates for BCFL and compare their performance  with the one with the optimal compression rate to evaluate how much performance can be improved through our analysis. By default, we set $k=d*1\%, d*2\%$ and $d*3\%$ for the second kind of baselines. In fact, we have $k=d $ for BFL  without model compression. Note that we let $k^*$ and $\lambda^*$ represent the optimal settings in BCFL obtained by solving $\mathbb{P}2$.


\subsubsection{Network Simulator} 
We refer to \cite{zhong2021}, \cite{tran2019federated} to simulate wireless communications between miners and clients (or miners) in our experiments. The uplink and downlink transmission rates  of clients and miners  are random variables obeying Gaussian distribution. The mean value is set according to Shannon's formula $\mu=BW\log_2(1+\frac{gP_t}{P_n})$, where $BW=20MHz$ is the channel bandwidth and $g=10^{-8}$ is the channel gain. Furthermore, we set the transmission power $P_t = 0.5W$ and the noise energy $P_n = 10^{-10}W$ to simulate the rate of wireless transmission to measure the time required for communications during training.  We set the same rate for all clients and miners for both downlink and uplink in our experiments. At each transmission, we take samples from the Gaussian distribution to simulate the fluctuation of the transmission rate in real networks. The mean and standard deviation of the Gaussian distribution are $\mu$ and $0.1\mu$, respectively. 
Based on the network speed model, we have $u_{\uparrow, i}=u_{\downarrow, i}=u_j=\mu$. 

For the network topology, we set up $50$ clients and $50 $ miners, \emph{i.e.,} $M=N$, by default unless we state otherwise. According to the previous work \cite{9399813}, we set up a one-to-one connection between miners and clients but all miners are fully connected with each other. Note that the values of $N$ and $M$ can be different in practice, and each miner can connect to multiple clients. It only slightly alters  coefficients in $h(k,\lambda)$ without affecting our analysis framework. 

By conducting local iterations in advance,  we can measure that  $\tau_{local}\approx 0.2s$ and $0.08s$ with CIFAR-10 and FEMNIST, respectively. But $\tau_{aggre}\approx 0$ because of the simplicity of the aggregation operation.

\subsubsection{Solving $\mathbb{P}2$}

In BCFL, it is required to solve $\mathbb{P}_2$ to determine $k^*$ and $\lambda^*$. Before solving $\mathbb{P}_2$, it is necessary to estimate a series of vital parameters in $h(k, \lambda)$ and the convergence rate. How to estimate these parameters are briefly described as below.

\begin{table}[h]
\centering
\caption{Estimation of parameters in the convergence rate.}
\begin{tabular}{|c|c|c|c|c|}
\hline
Scenario & $L$ & $G^2$ & $\Gamma_G^2$ & $F(\mathbf{w}_0)-F^*$ \\ 
\hline
CIFAR-10+IID & 0.45 & 0.15 & 0.00044 & 2.30 \\
\hline
CIFAR-10+non-IID & 0.16 & 10.44 & 0.029 & 2.19 \\
\hline
FEMNIST & 6.49 & 1.17 & 0.0035 & 4.08 \\
\hline
\end{tabular}
\label{parameterValue}
\end{table}

In the first round of global iteration, we arbitrarily set $k=1\%d$ for CIFAR-10 and $k=0.5\%d$ for FEMNIST to obtain 
$\mathbf{w}_0$ (initial parameters randomly generated) and $\mathbf{w}_E$. Then, $L$ in Assumption~\ref{Assump:ConSmoo} can be estimated based on $\mathbf{w}_0$  and $\mathbf{w}_E$. We use the loss function value with input $\mathbf{w}_E$ as its approximation of $\mathbb{E}[F(\mathbf{w}_{0})]- F^*$ by assuming that $F^*\approx 0$. In our experiments, the full-batch of local samples will be used for local iterations, and thus $\sigma_i=0$.\footnote{If a mini-batch is used for local iterations, $\sigma_i$ is can be estimated locally}.  Meanwhile, clients are also required to upload the  norm of their gradients to miners in the first global iteration so that we can estimate $G$.  Based on the above method , we can estimate the values of these parameters in the convergence rate which are listed in Table.~\ref{parameterValue}.

According to the expression of $h(k, \lambda)$ in Eq.~\eqref{EQ:TimeCost}, we stil need to estimate $\tau_{local}$ and $\tau_{aggre}$ before we can solve  $\mathbb{P}2$. By executing local iterations in the first global iteration,  we can easily measure that  $\tau_{local}\approx 0.2s$ and $0.08s$ with CIFAR-10 and FEMNIST, respectively, on our experiment platform. But $\tau_{aggre}\approx 0$ because of the simplicity of the aggregation operation. 

With these estimated parameters in hand, we can finally solve $\mathbb{P}2$ to deduce $k^*$ and $\lambda^*$ for our experiments. Their values are also presented together with each experiment in the next subsection.

\subsection{Experimental Result}

\subsubsection{Comparison of Model Accuracy}
%
%

\begin{figure*}[h]
\centering
\includegraphics[width=\linewidth]{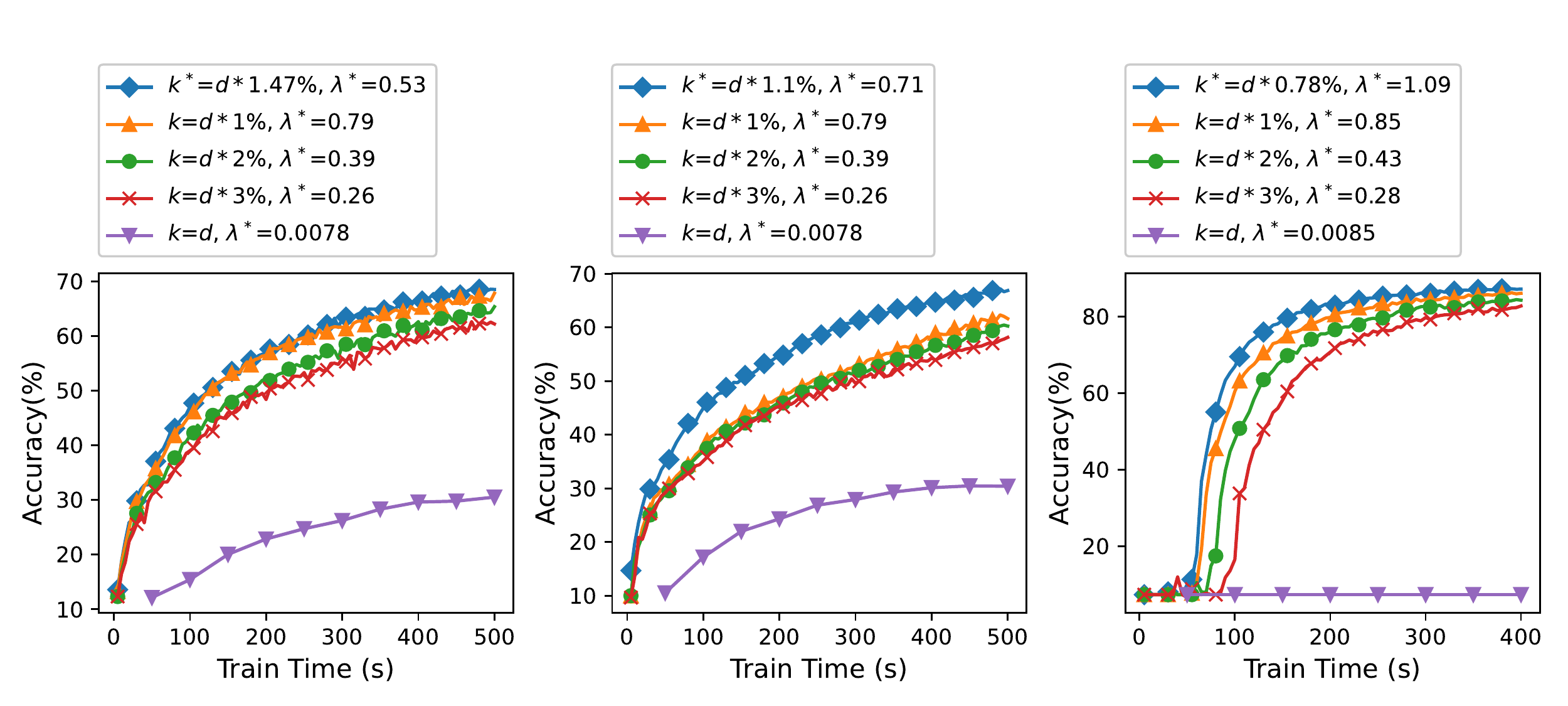}
\caption{ {Comparison of accuracy between  BFL and BCFL with different  compression rates  under IID+CIFAR-10 (left), non-IID+CIFAR-10 (middle) and FEMNIST (right)}. }
\label{Accuracy}
\end{figure*}
We compare the model accuracy of BCFL with  baselines within a fixed training time span $Y=500s$ and $400s$ on CIFAR-10 and FEMNIST, respectively. 
The results are plotted in Fig.~\ref{Accuracy}, where the x-axis represents the elapsed training time and the y-axis represents the model accuracy on the test set. From experiment results in Fig.~\ref{Accuracy}, we can observe that: 
\begin{itemize}
\item Compressing model updates in BFL can effectively improve the training efficiency. As long as $k$ is  an arbitrary number much  smaller than $d$, BCFL can achieve much higher model accuracy than  BFL under all experimental scenarios. Apparently, the reason is that the heavy communication load in BFL consumes excessive training time. 
\item Setting $k^*$ and $\lambda^*$ in BCFL achieves the highest model accuracy than all other baselines. In particular,  for the non-IID+CIFAR-10 case, the improvement is significant. 
\item Since FEMNIST has 62 labels, which is much more than that of CIFAR-10, more global iterations are required to train an effective model. Therefore, the model trained in the given time span has poor performance for BFL without compressing communications.
\end{itemize}

\subsubsection{Comparison of Model Accuracy with Fixed $\lambda$}
\begin{figure*}[h]
\centering
\includegraphics[width=\linewidth]{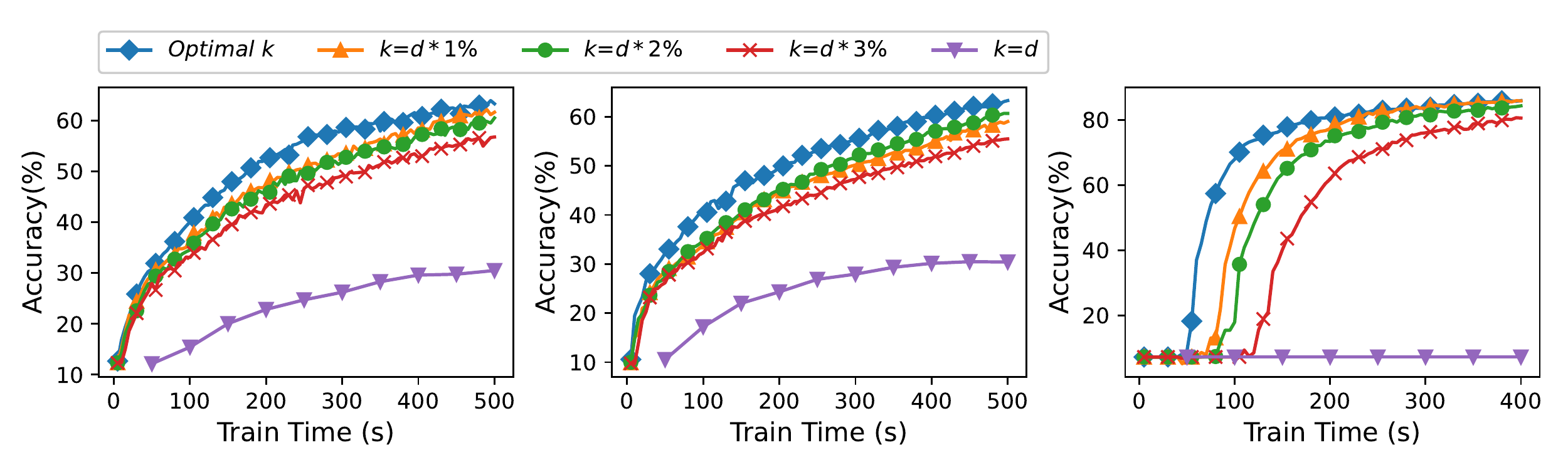}
\caption{ {Comparison of accuracy between BFL and BCFL with different compression rates and fixed $\lambda=0.4$ under IID+CIFAR-10 (left), non-IID+CIFAR-10 (middle) and FEMNIST (right).} }
\label{FixLamAccu}
\end{figure*}

In BCFL, $\lambda$ is a parameter to control the rate for block generation. It is possible to fix $\lambda$ as any number to control the block generation rate. 
We next compare the performance of BCFL with baselines by arbitrarily fixing $\lambda=0.4$ in Fig.~\ref{FixLamAccu}.
In fact, $\mathbb{P}2$ degenerates to a convex optimization problem when $\lambda$ is fixed in advance and it is expected that BCFL can still achieve the best performance.


We keep other parameters the same as these in the previous experiment. 
From the experimental results presented in Fig.~\ref{FixLamAccu}, we can also observe that 
the model accuracy of BCFL is always better than all other baselines. Note that the improvement of BCFL  under the non-IID+CIFAR-10 case  is not as significant as that in the previous experiment. The reason is that $\lambda$ is a fixed number, not assigned with its optimal value.

\subsubsection{Varying Compression Rates}
\begin{figure*}[h]
\centering
\includegraphics[width=\linewidth]{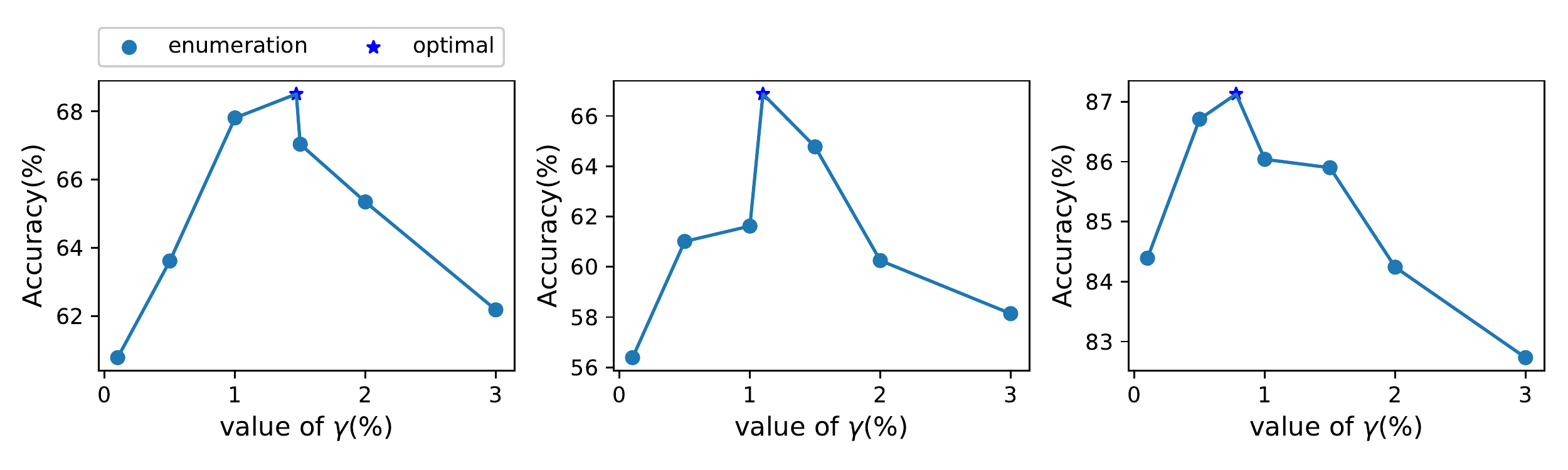}
\caption{ {Comparison of final model accuracy of BCFL by enumerating different compression rates under IID+CIFAR-10 (left), non-IID+CIFAR-10 (middle) and  FEMNIST (right)} }
\label{DifferentK}
\end{figure*}
To further validate that BCFL can optimally set the compression rate to maximize the final model accuracy, we conduct more experiments by enumerating different compression rates, \emph{i.e.}, setting  $k = 0.1\%*d, 0.5\%*d, 1\%*d, 1.5\%*d, 2\%*d$ and $3\%*d,$ respectively in BCFL. The training time span $Y$ is fixed as  500s, 500s, 400s for three experiments. For these experiments, $\lambda$ is set as its optimal value once $k$ is fixed. 


Experiment results are shown in
Fig.~\ref{DifferentK}, in which the x-axis represents the compression rate and the y-axis represents the final model accuracy after $Y$ training time. 
The results in  Fig.~\ref{DifferentK} manifest that our theoretical optimal setting can achieve the highest final model accuracy among all baselines with  enumerated compression rates. It indicates the effectiveness of our optimization analysis.\footnote{Due to limited computation capacity and the large value of $d$, we cannot enumerate all possible $k$'s in our experiments.}  


\subsubsection{Comparison of Training Time Consumption}

\begin{table}[h]
\centering
\caption{Comparison of different algorithms with IID distribution and 61\% model accuracy in CIFAR-10.}
\begin{tabular}{|c|c|c|c|c|}
\hline
& \tabincell{c}{Comm.\\Rate} & Comm. Traffic & Train Time & \tabincell{c}{train time\\reduced to}\\
\hline
BFL & 1 & 39021.31MB & 4953.46s & $100\%$\\
\hline
\textbf{\tabincell{c}{$k^*$}} & \textbf{44.43} & \textbf{867.80MB} & \textbf{260.12s} & \textbf{5.25\%}\\
\hline
$k=1\%*d$ & 65.31 & 648.54MB & 285.19s & $5.76\%$\\
\hline
$k=2\%*d$ & 32.65 & 1165.54MB & 380.26s & $7.68\%$\\
\hline
$k=3\%*d$ & 21.77 & 1673.93MB & 436.56s & $8.81\%$\\
\hline
\end{tabular}
\label{IIDCommunicationCIFAR}
\end{table}

\begin{table}[h]
\centering
\caption{Comparison of different algorithms with non-IID distribution and 58\% model accuracy in CIFAR-10.}
\begin{tabular}{|c|c|c|c|c|}
\hline
& \tabincell{c}{Comm.\\Rate} & Comm. Traffic & Train Time & \tabincell{c}{train time\\reduced to}\\
\hline
BFL & 1 & 35148.92MB & 4650.21s & $100\%$\\
\hline
\textbf{\tabincell{c}{$k^*$}} & \textbf{59.37} & \textbf{686.81MB} & \textbf{245.11s} & \textbf{5.27\%}\\
\hline
$k=1\%*d$ & 65.31 & 628.07MB & 395.87s & $8.51\%$\\
\hline
$k=2\%*d$ & 32.65 & 1184.25MB & 450.59s & $9.69\%$\\
\hline
$k=3\%*d$ & 21.77 & 1673.93MB & 501.51s & $10.78\%$\\
\hline
\end{tabular}
\label{non-IIDCommunicationCIFAR}
\end{table}

\begin{table}[h]
\centering
\caption{Comparison of different algorithms with non-IID distribution and 82\% model accuracy in FEMNIST.}
\begin{tabular}{|c|c|c|c|c|}
\hline
& \tabincell{c}{Comm.\\Rate} & Comm. Traffic & Train Time & \tabincell{c}{train time\\reduced to}\\
\hline
BFL & 1 & 36460.87MB & 4553.17s & $100\%$\\
\hline
\textbf{\tabincell{c}{$k^*$}} & \textbf{85.73} & \textbf{513.52MB} & \textbf{185.42s} & \textbf{4.07\%}\\
\hline
$k=1\%*d$ & 65.31 & 605.22MB & 230.58s & $5.06\%$\\
\hline
$k=2\%*d$ & 32.65 & 1138.26MB & 290.09s & $6.37\%$\\
\hline
$k=3\%*d$ & 21.77 & 1668.04MB & 355.04s & $7.80\%$\\
\hline
\end{tabular}
\label{non-IIDCommunicationFEMNIST}
\end{table}

To further explore the underling reason how BCFL improves the model accuracy, we compare the consumed training time of each algorithm to reach the target model accuracy by fixing the target model accuracy as 61\%, 58\% and 82\%  for three experiment scenarios, respectively. 
We keep other settings the same as these in the experiment in Fig.~\ref{Accuracy}. The comparisons of training time  are shown in Tables ~\ref{IIDCommunicationCIFAR} (for CIFAR-10+IID), ~\ref{non-IIDCommunicationCIFAR} (for CIFAR-10+non-IID) and ~\ref{non-IIDCommunicationFEMNIST} (for FEMNIST), respectively. Here BFL represents the blockchain-based FL without model compression. 


From results shown in these tables, we can draw the following conclusions.
\begin{itemize}
\item The BCFL algorithm can always reach the target final model accuracy with the shortest training time. Despite that BCFL  may not be the one with the least amount of communication traffic, it takes a smaller number of global iterations to complete model training such that the total training time cost is minimized. More importantly,   BCFL can reduce the consumed training time by more than 90\% compared with BFL.

\item Our results show the merit of our algorithm to optimally adjust the compression rate in BCFL. If the compression rate is too high, it can  considerably slow down the convergence rate presented in Theorem~\ref{ConvergenceOfModel}.  Although the communication traffic can reduced significantly, it takes more global iterations to reach the target model accuracy, Inversely, if the compression rate is too low, it consumes  excessive communication traffic to complete a round of global iteration, which inevitably consumes  exorbitant training time.

\end{itemize}

\subsubsection{Varying Client and Miner Population}
\begin{figure*}[h]
\centering
\includegraphics[width=\linewidth]{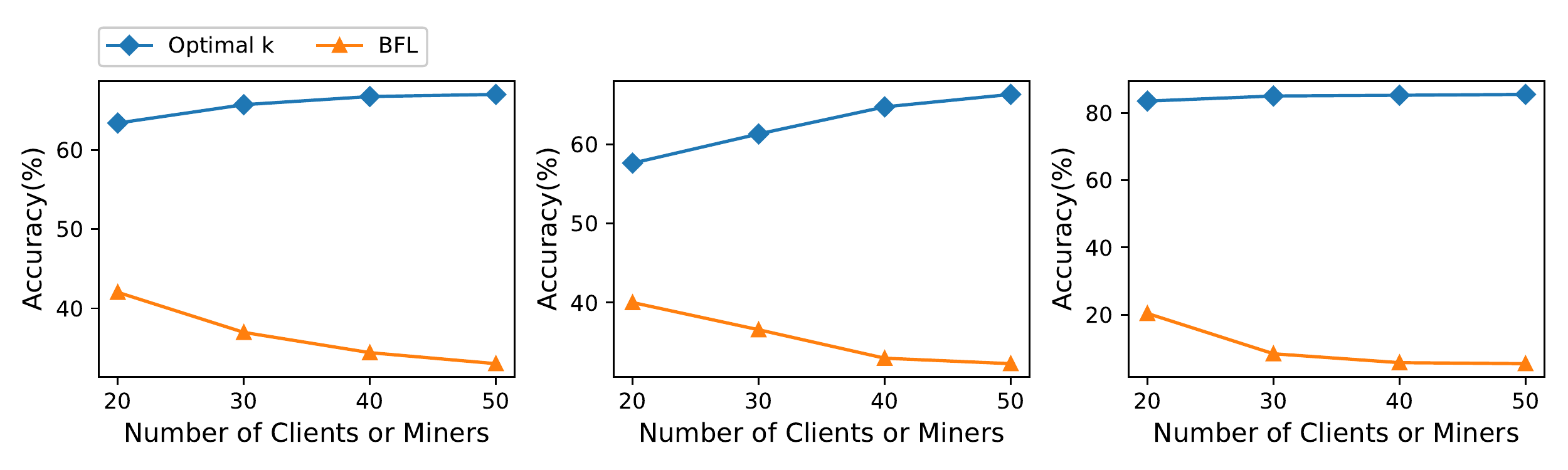}
\caption{ {Comparison of accuracy between BFL and BCFL by varying dynamic client and miner populations under  CIFAR-10+IID (left), CIFAR-10+non-IID (middle) and  FEMNIST (right)} }
\label{DynamicClient}
\end{figure*}

In this experiment, we change the system scale by setting the number of clients and miners as $N = M = 20, 30, 40$ or $50$, respectively, to investigate the influence of the system scale on model accuracy performance.  Meanwhile, to evaluate the robustness of BCFL, we consider the dynamics of networking by assuming that each client has 10\% probability not to return model updates due to the sudden changes of network conditions such as network congestion or failure in each global iteration. 
When we the change client population,  the distribution of CIFAR-10 samples on clients is modified accordingly.  For example, if the number of clients is doubled, the number of  samples on each client is reduced by 50\%. For FEMNIST, the samples on each client are from  a particular writer. Therefore, as the number of clients decreases, the number of samples on each client is unchanged, but the total number of samples in the system becomes smaller.


Experiment results  are shown in Fig.~\ref{DynamicClient}, where the x-axis represents the number of clients, and the y-axis represents the final trained model performance. 
The training time spans used for three different data distribution scenarios are $Y=500s, 500s$ and $400s$, respectively.
By inspecting the experiment results, we can conclude that
\begin{itemize}
\item Experiment results  show the robustness of BCFL. Even if the network is dynamic with occasionally offline clients,  BCFL can still achieve the higher model training  performance than that of BFL. 
\item The model accuracy of BCFL  under the first two scenarios becomes higher as the number of clients increases. The reason lies in that more clients can bring more computation capacity to the system. As the number of samples allocated to each client is reduced, clients can complete local iterations faster. Consequently, the final model accuracy is improved by conducting more global iterations within a fixed training time span.  
\item The increase of model accuracy of BCFL with the increase of client population cannot be observed in the experiment with FEMNIST. The reason is that the sample population on each client is unchanged as the increase of client population, and thereby the local training time cannot be reduced. 
\item It is worthy noting that the model accuracy of BFL becomes worse with the increase of client population under all three experiment scenarios. The reason is that BFL cannot effectively prohibit the increase of communication traffic. As more clients reside in the system, more communication traffic is generated that can seriously prolong the communication time resulting in a lower model accuracy at last. 
\end{itemize}

\section{Conclusion}\label{Conclusion}

Combing blockchain with federated learning is of essential importance to construct a decentralized, auditable and trustworthy federated learning system. Nonetheless, the huge communication traffic generated by BFL has severely hinder the application of BFL in reality. Our work makes a significant attempt trying to solve this critical issue by compressing model updates in BFL. We proposed the novel BCFL framework, and proved its convergence by leveraging the $Top_k$ compression algorithm under non-IID sample distribution and non-convex loss. Based on the derived convergence rate, we further formulate the optimization problem to maximize the final model accuracy with respect to compression rate and block generation rate. The problem is a bi-convex optimization problem, which can be solved efficiently.  
Lastly, we conduct extensive experiments with public datasets to not only validate the correctness of our  analysis but also demonstrate the notable performance of our BCFL algorithm. In particular, by compressing model updates, our communication-efficient framework can reduce the training time by about 95\% without compromising model accuracy. Our future work is to implement the prototype of BCFL and investigate its performance in more practical environment. 


\appendices
\section{Notation List}\label{notation}
To ease the interpretation of our analysis, we list  brief explanations of major notations used in the time cost analysis  and convergence analysis of Theorem~\ref{ConvergenceOfModel} in Table~\ref{NotationListOfTau} and Table~\ref{NotationListOfTheorem}, respectively.

\begin{table}[h]
\centering
\caption{Notation Meaning in Time Cost Analysis}
\begin{tabular}{|m{1cm}<{\centering}|m{6cm}<{\centering}|}
\hline
Notation & Meaning \\
\hline
$Y$ & total training time \\
\hline
$\tau_{local}$ & time required for clients to train the model locally\\
\hline
$\tau_{\uparrow, i}$ & time required for client $i$ to upload the model updates to corresponding miner \\
\hline
$\tau_{cross, j}$ & time required for miner $j$ to receive the transactions crossed from other miners \\
\hline
$\tau_{mine, j}$ &  time required for miner $j$ to complete the consensus mechanism \\
\hline
$\tau_{pro}$ & time required for winning miner to propagate the block \\
\hline
$\tau_{\downarrow, i}$ & time required for client $i$ to download the latest block from miner\\
\hline
$\tau_{aggre}$ & time required for clients to aggregate the model updates in blocks \\
\hline
\end{tabular}
\label{NotationListOfTau}
\end{table}

\begin{table}[h]
\centering
\caption{Notation Meaning in Convergence Analysis}
\begin{tabular}{|m{1cm}<{\centering}|m{6cm}<{\centering}|}
\hline
Notation & Meaning \\
\hline
$\mathbf{z}_T$ & the gradients of a model extracted with a probability of $\frac{1}{NT}$ \\
\hline
$E$ & number of local epochs on the clients \\
\hline
$F^*$ & optimal value of global loss function \\
\hline
$C$ & a constant satisfying $\frac{C}{\sqrt{T}}\le\frac{1}{16L}$ \\
\hline
$L$ & the loss functions satisfy $L-smooth$ \\
\hline
$\Gamma_G^2$ & quantification of non-IID \\
\hline
$b$ & batch data size \\
\hline
$N$ & number of clients \\
\hline
$\sigma_i$ & upper bound on variance of stochastic gradient \\
\hline
$R$ & number of global iterations \\
\hline
\end{tabular}
\label{NotationListOfTheorem}
\end{table}

\section{Proof of Theorem~\ref{ConvergenceOfModel}} \label{ProofOfConvergence}
To analyze the convergence of  BCFL, we first construct an auxiliary sequence as follow,
$$\widetilde{\mathbf{w}}^i_{t+1}=\widetilde{\mathbf{w}}^i_{t}-\eta\nabla F_{i}(\mathbf{w}^i_t,\mathcal{B}^i_t);\quad\widetilde{\mathbf{w}}_{t+1}=\frac{1}{N}\sum_i\widetilde{\mathbf{w}}^i_{t+1},$$
where $\widetilde{\mathbf{w}}^i_0=\mathbf{w}^i_0$. Further, we define $\widehat{\mathbf{w}}_t=\frac{1}{N}\sum_i\mathbf{w}_t^i$. Accordingly, $\{\widetilde{\mathbf{w}}_{t}, \forall t\}$ represents centralized sequence trained by uncompressed model updates, while $\{\widehat{\mathbf{w}}_t, \forall t\}$ represents centralized sequence trained by compressed model updates.
\subsection{Key Lemmas}
To derive the convergence of the model, we leverage the following lemmas.
\begin{lemma}
If the model will be updated $E$ epochs in clients in each global iteration during training process, then the following inequality holds for any $i$ and $t$,
$$\mathbb{E}[\|\mathbf{m}^i_t\|^2]\leq4\frac{\eta^2(1-\gamma)^2}{\gamma^2}E^2G^2.$$
\end{lemma}

\begin{lemma}
According to the  definition of $\widehat{\mathbf{w}}_t$, we can get
$$\widehat{\mathbf{w}}_t-\widetilde{\mathbf{w}}_t=\frac{1}{N}\sum_i \mathbf{m}^i_t.$$
\end{lemma}

\begin{lemma}
Assume that the number of local training epochs of the clients is $E$, we can get
$$\frac{1}{N}\sum_i\mathbb{E}\|\widehat{\mathbf{w}}_t-\mathbf{w}^i_t\|^2\leq\eta^2G^2E^2.$$
\end{lemma}

The above lemmas have been proved in \cite{basu2020qsparse}, which will not be repeated in this paper.

\subsection{Convergence of the Algorithm}
According to \cite{basu2020qsparse}, we can get
\begin{equation}
\begin{aligned}
&F(\widetilde{\mathbf{w}}_{t+1})-F(\widetilde{\mathbf{w}}_{t})\\
&\leq-\eta \left<\nabla F(\widetilde{\mathbf{w}}_{t}),\frac{1}{N}\sum_i\nabla F_i(\mathbf{w}^i_{t},\mathcal{B}^i_t)\right>\\
&\quad+\eta^2L\|\frac{1}{N}\sum_i\nabla F_{i}(\mathbf{w}^i_t)\|^2+\eta^2L\|\mathbf{p}_t-\bar{\mathbf{p}}_t\|^2,
\end{aligned}
\end{equation}
where $\mathbf{p}_t=\frac{1}{N}\sum_i \nabla F_i(\mathbf{w}^i_t,\mathcal{B}^i_t),\ \bar{\mathbf{p}}_t=\frac{1}{N}\sum_i \nabla F_i(\mathbf{w}^i_t)$ representing average gradients.
To solve the expected value of the stochastic gradient sampling in the $t$-th round and $<\mathbf{a}, \mathbf{b}>=\frac{1}{2}(\|\mathbf{a}\|^2+\|\mathbf{b}\|^2-\|\mathbf{a}-\mathbf{b}\|^2)$, we can get
\begin{eqnarray}
&&\mathbb{E}F(\widetilde{\mathbf{w}}_{t+1})-F(\widetilde{\mathbf{w}}_{t})\notag\\
&&\leq-\frac{\eta}{2}\Big(\|\nabla F(\widetilde{\mathbf{w}}_{t})\|^2+\|\frac{1}{N}\sum_i \nabla F_i(\mathbf{w}^i_{t})\|^2\notag\\
&&\quad-\|\nabla F(\widetilde{\mathbf{w}}_{t})-\frac{1}{N}\sum_i \nabla F_i(\mathbf{w}^i_{t})\|^2\Big)\notag\\
&&\quad +\eta^2L\|\frac{1}{N}\sum_i\nabla F_i(\mathbf{w}^i_t)\|^2+\frac{\eta^2L}{bN^2}\sum_i\sigma_i^2\notag\\
&&=-\frac{\eta}{2}\Big( \|\nabla F(\widetilde{\mathbf{w}}_{t})\|^2+\|\frac{1}{N}\sum_i \nabla F_i(\mathbf{w}^i_{t})\|^2\notag\\
&&\quad-\Big\|\frac{1}{N}\sum_i \nabla F_i(\widetilde{\mathbf{w}}_{t})-\frac{1}{N}\sum_i \nabla F_i(\mathbf{w}^i_{t})\Big\|^2\Big)\notag\\
&&\quad +\eta^2L\|\frac{1}{N}\sum_i\nabla F_i(\mathbf{w}^i_t)\|^2+\frac{\eta^2L}{bN^2}\sum_i\sigma_i^2\notag\\
&&\le\frac{\eta^2L}{bN^2}\sum_i\sigma_i^2-\frac{\eta}{2}\Big( \|\nabla F(\widetilde{\mathbf{w}}_{t})\|^2+\|\frac{1}{N}\sum_i \nabla F_i(\mathbf{w}^i_{t})\|^2\notag\\
&&\quad-\frac{1}{N}\sum_iL^2\|\widetilde{\mathbf{w}}_{t}-\mathbf{w}^i_{t}\|^2\Big)+\eta^2L\|\frac{1}{N}\sum_i\nabla F_i(\mathbf{w}^i_t)\|^2\notag\\
&&= -\frac{\eta}{2N}\sum_i \left(\|\nabla F(\widetilde{\mathbf{w}}_{t})\|^2-L^2\|\widetilde{\mathbf{w}}_{t}-\mathbf{w}^i_{t}\|^2\right)\notag\\
&&\quad +\frac{2\eta^2L-\eta}{2}\|\frac{1}{N}\sum_i\nabla F_i(\mathbf{w}^i_t)\|^2+\frac{\eta^2L}{bN^2}\sum_i\sigma_i^2\notag\\
&&\le -\frac{\eta}{2N}\sum_i \left(\|\nabla F(\widetilde{\mathbf{w}}_{t})\|^2-L^2\|\widetilde{\mathbf{w}}_{t}-\mathbf{w}^i_{t}\|^2\right)\notag\\
&&\quad +\frac{2\eta^2L}{2}\|\frac{1}{N}\sum_i\nabla F_i(\mathbf{w}^i_t)\|^2+\frac{\eta^2L}{bN^2}\sum_i\sigma_i^2\notag\\
&&\leq \frac{2\eta^2L}{2N}\sum_i\|\nabla F_i(\mathbf{w}^i_t)-\nabla F(\mathbf{w}^i_t) + \nabla F({\mathbf{w}}^i_t)\|^2\notag\\
&& \quad -\frac{\eta}{2N}\sum_i \left(\|\nabla F(\widetilde{\mathbf{w}}_{t})\|^2-L^2\|\widetilde{\mathbf{w}}_{t}-\mathbf{w}^i_{t}\|^2\right)\notag\\
&&\quad+\frac{\eta^2L}{bN^2}\sum_i\sigma_i^2\notag\\
&&\leq \frac{2\eta^2L}{N}\sum_i\|\nabla F_i(\mathbf{w}^i_t)-\nabla F(\mathbf{w}^i_t)\|^2\notag\\
&&\quad+\frac{2\eta^2L}{N}\sum_i\|\nabla F(\mathbf{w}^i_t)\|^2+\frac{\eta^2L}{bN^2}\sum_i\sigma_i^2\notag\\
&& \quad -\frac{\eta}{2N}\sum_i \left(\|\nabla F(\widetilde{\mathbf{w}}_{t})\|^2-L^2\|\widetilde{\mathbf{w}}_{t}-\mathbf{w}^i_{t}\|^2\right)\notag\\
&&\le\frac{2\eta^2L}{N}\sum_i\|\nabla F(\mathbf{w}^i_t)\|^2+2\eta^2L\Gamma_G^2+\frac{\eta^2L}{bN^2}\sum_i\sigma_i^2\notag\\
&&\quad-\frac{\eta}{2N}\sum_i \left(\|\nabla F(\widetilde{\mathbf{w}}_{t})\|^2-L^2\|\widetilde{\mathbf{w}}_{t}-\mathbf{w}^i_{t}\|^2\right)\notag\\
&&=\frac{2\eta^2L}{N}\sum_i\|\nabla F({\mathbf{w}}^i_t)\|^2+2\eta^2L\Gamma_G^2\notag\\
&&\quad-\frac{\eta}{2N}\sum_i \left(\|\nabla F(\widetilde{\mathbf{w}}_{t})\|^2+L^2\|\widetilde{\mathbf{w}}_{t}-\mathbf{w}^i_{t}\|^2\right)\notag\\
&&\qquad +\frac{\eta^2L}{bN^2}\sum_i\sigma_i^2+\frac{\eta L^2}{N}\sum_i\|\widetilde{\mathbf{w}}_{t}-\mathbf{w}^i_{t}\|^2.\notag
\end{eqnarray}

The term $\|\nabla F({\mathbf{w}}^i_t)\|^2$ can be bounded as below. 
\begin{equation}
\begin{aligned}
\|\nabla F(\mathbf{w}^i_t)\|^2&\leq2\|\nabla F(\mathbf{w}^i_t)-\nabla F(\widetilde{\mathbf{w}}_t)\|^2+2\|\nabla F(\widetilde{\mathbf{w}}_t)\|^2\\
&\leq 2L^2\|\mathbf{w}^i_t-\widetilde{\mathbf{w}}_t\|^2+2\|\nabla F(\widetilde{\mathbf{w}}_t)\|^2.
\end{aligned}
\end{equation}
If $\eta\le\frac{1}{16L}$, we can get
\begin{equation}
\begin{aligned}
&\frac{\eta}{8N}\sum_i\|\nabla F(\mathbf{w}^i_t)\|^2\leq F(\widetilde{\mathbf{w}}_{t})- \mathbb{E}F(\widetilde{\mathbf{w}}_{t+1})\\
&+2\eta^2L\Gamma_G^2+\frac{\eta^2L}{bN^2}\sum_i\sigma_i^2+\frac{\eta L^2}{N}\sum_i\|\widetilde{\mathbf{w}}_{t}-\mathbf{w}^i_{t}\|^2.
\end{aligned}
\end{equation}
By taking the expected value of each item in the above formula, we can get
\begin{equation}
\begin{aligned}
&\frac{\eta}{8N}\sum_i\mathbb{E}\|\nabla F(\mathbf{w}^i_t)\|^2\leq \mathbb{E}[F(\widetilde{\mathbf{w}}_{t})]- \mathbb{E}[F(\widetilde{\mathbf{w}}_{t+1})]\\
&\quad+2\eta^2L\Gamma_G^2+\frac{\eta^2L}{bN^2}\sum_i\sigma_i^2+2\eta L^2\mathbb{E}\|\widetilde{\mathbf{w}}_{t}-\widehat{\mathbf{w}}_{t}\|^2\\
&\quad+\frac{2\eta L^2}{N}\sum_i\mathbb{E}\|\widehat{\mathbf{w}}_{t}-\mathbf{w}^i_{t}\|^2.
\end{aligned}
\end{equation}
According to Lemmas 1 and 2, we can get $\mathbb{E}\|\widehat{\mathbf{w}}_t-\widetilde{\mathbf{w}}_{t}\|^2\le\frac{4\eta^2(1-\gamma^2)}{\gamma^2}G^2E^2$. Therefore, we can leverage Lemmas 1-3 to derive \begin{equation}
\begin{aligned}
&\frac{\eta}{8N}\sum_i\mathbb{E}\|\nabla F(\mathbf{w}^i_t)\|^2\leq \mathbb{E}[F(\widetilde{\mathbf{w}}_{t})]- \mathbb{E}[F(\widetilde{\mathbf{w}}_{t+1})]\\
&\quad+2\eta^2L\Gamma_G^2+\frac{\eta^2L}{bN^2}\sum_i\sigma_i^2+\frac{8\eta^3(1-\gamma^2)}{\gamma^2}L^2G^2E^2\\
&\quad+2\eta^3 L^2G^2E^2.
\end{aligned}
\end{equation}
By accumulating the above inequality from $t=0$ to $T-1$ and dividing $T$ on both sides, we can get
\begin{equation}
\begin{aligned}
&\frac{1}{8TN}\sum_t\sum_i\mathbb{E}\|\nabla F(\mathbf{w}^i_t)\|^2\leq \frac{\mathbb{E}[F(\widetilde{\mathbf{w}}_{0})]- F^*}{\eta T}\\
&\quad+2\eta L\Gamma_G^2+\frac{\eta L}{bN^2}\sum_i\sigma_i^2+\frac{8\eta^2(1-\gamma^2)}{\gamma^2}L^2G^2E^2\\
&\quad+2\eta^2 L^2G^2E^2.
\end{aligned}
\end{equation}
By setting $\eta=\frac{C}{\sqrt{T}}$ and substituting into the above equation, we can prove Theorem~\ref{ConvergenceOfModel}.

\section{Proof of Theorem~\ref{biConvex}} \label{ProofofBiConvex}
\begin{proof}
According to the content in \cite{pokhrel2020federated}, we can know that when $k$ is fixed, $h(k, \lambda)$ is a convex function with respect to $\lambda$. At this time, the objective function $\mathcal{J}^2(k, \lambda)$ is also convex. Therefore we next discuss the properties of $\mathcal{J}^2(k, \lambda)$ when $\lambda$ is fixed.

To simplify the analysis process, we denote $\omega = s+\frac{\lceil\log_2d\rceil}{8}$ as the size of each compressed model update consisting of value and position ID. We define
$$\Lambda_T=\max_{i \in \mathcal{N}}\frac{\omega}{u_{\uparrow,i}}+\max_{j \in \mathcal{M}}\frac{(N-N_j)\omega}{u_j}+\max_{i \in \mathcal{N}}\frac{N\omega}{u_{\downarrow,i}},$$
$$\ \Lambda_P=\max_{j \in \mathcal{M}/j_w}\frac{N\omega}{u_j},\Lambda_F=\lambda\sum_{j\in\mathcal{M}/j_w}\frac{N\omega}{u_j},$$
for parameters whose values are constant.
Thence, $h(k, \lambda)=\tau_{local}+\tau_{aggre}+\Lambda_Tk+\frac{1}{\lambda}\exp(\Lambda_Fk)+\Lambda_Pk\exp(\Lambda_Fk)$. It is obvious that the first term in the objective function $\mathcal{J}^2(k, \lambda)$ is convex when $\lambda$ is fixed. So we will focus on the convexity of second term.
\begin{equation}
\label{EQ:ConvexOfF}
\begin{aligned}
&\frac{h(k, \lambda)}{k^2}=(\tau_{local}+\tau_{aggre})k^{-2}+\Lambda_Tk^{-1}\\
&\quad+\frac{\frac{1}{\lambda}\exp(\Lambda_Fk)}{k^2}+\frac{\Lambda_P\exp(\Lambda_Fk)}{k}.
\end{aligned}
\end{equation}  

Both the first and second terms in Eq.~\eqref{EQ:ConvexOfF} are convex, thus we discuss the convexity of the third term first. We can get
\begin{equation}
\begin{aligned}
&\left(\frac{\frac{1}{\lambda}\exp(\Lambda_Fk)}{k^2} \right)^{'}\\
&=\frac{1}{\lambda}\Lambda_Fk^{-2}\exp(\Lambda_Fk)-2\frac{1}{\lambda}k^{-3} \exp(\Lambda_Fk).
\end{aligned}
\end{equation}
According to the result of the above formula, we can further deduce that
\begin{equation}
\begin{aligned}
&\left(\frac{\frac{1}{\lambda}\exp(\Lambda_Fk)}{k^2} \right)^{''}=-2\frac{1}{\lambda}\Lambda_Fk^{-3}\exp(\Lambda_Fk)\\
&\quad+\frac{1}{\lambda}\Lambda_F^2k^{-2}\exp(\Lambda_Fk) +6\frac{1}{\lambda}k^{-4}\exp(\Lambda_Fk)\\
&\quad-2\frac{1}{\lambda}\Lambda_Fk^{-3}\exp(\Lambda_Fk)\\
&=\frac{1}{\lambda}\Lambda_Fk^{-4}\exp(\Lambda_Fk)\left(-4k+\Lambda_Fk^2+\frac{6}{\Lambda_F} \right).
\end{aligned}
\end{equation}
Considering $\Lambda_F > 0$ and $(-4)^2-4*\Lambda_F*\frac{6}{\Lambda_F}<0$, so $\left(\frac{\frac{1}{\lambda}\exp(\Lambda_Fk)}{k^2} \right)^{''}>0$, which means that the third term in Eq.~\eqref{EQ:ConvexOfF} is convex. 
Therefore, $\frac{h(k, \lambda)}{k^2}$ in the objective function $\mathcal{J}^2(k, \lambda)$ is convex when $\lambda$ is fixed since we can prove that $\frac{\Lambda_P\exp(\Lambda_Fk)}{k}$ is convex in the same way.

Identically, we can prove that $\frac{h(k, \lambda)}{k^{\frac{4}{3}}}$ in the objective function $\mathcal{J}^2(k, \lambda)$ is convex when $\lambda$ is fixed.

Through the above analysis, Theorem~\ref{biConvex} is proved.
\end{proof}

\ifCLASSOPTIONcaptionsoff
\newpage
\fi

%

\balance
\bibliographystyle{IEEEtran} 
\bibliography{reference}

\begin{IEEEbiography}[{\includegraphics[width=1in,height=1.25in,clip,keepaspectratio]{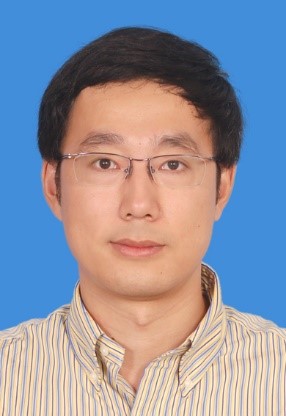}}]{Laizhong Cui}
is currently a Professor in the College of Computer Science and Software Engineering at Shenzhen University, China. He received the B.S. degree from Jilin University, Changchun, China, in 2007 and Ph.D. degree in computer science and technology from Tsinghua University, Beijing, China, in 2012. His research interests include Future Internet Architecture and Protocols, Edge Computing, Multimedia Systems and Applications, Blockchain, Internet of Things, Cloud and Big Data Computing, Computational Intelligence and Machine Learning. He led more than 10 scientific research projects, including National Key Research and Development Plan of China, National Natural Science Foundation of China, Guangdong Natural Science Foundation of China and Shenzhen Basic Research Plan. He has published more than 100 papers, including IEEE JSAC, IEEE TC, IEE TKDE, IEEE TMM, IEEE IoT Journal, IEEE TII, IEEE TVT, IEEE TNSM, ACM TOIT, IEEE TCBB, IEEE Network, IEEE INFOCOM, ACM MM, etc. He serves as an Associate Editor or a Member of Editorial Board for several international journals, including IEEE IoT Journal, IEEE Transactions on Network and Service Management, and International Journal of Machine Learning and Cybernetics. He is a Senior Member of the IEEE, and a Senior Member of the CCF.
\end{IEEEbiography}

\begin{IEEEbiography}[{\includegraphics[width=1in,height=1.25in,clip,keepaspectratio]{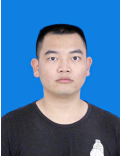}}]{Xiaoxin Su}
received his bachelor's degree from Shenzhen University in 2020. He is studying for a master's degree at Shenzhen University. His research interests include Federated Learning and Edge Computing.
\end{IEEEbiography}

\begin{IEEEbiography}[{\includegraphics[width=1in,height=1.25in,clip,keepaspectratio]{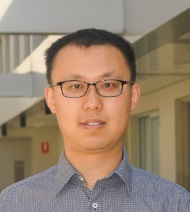}}]{Yipeng Zhou}
is a senior lecturer in computer science with School of Computing at Macquarie University, and the recipient of ARC DECRA in 2018. From Aug. 2016 to Feb. 2018, he was a research fellow with Institute for Telecommunications Research (ITR) of University of South Australia. From 2013.9-2016.9, He was a lecturer with College of Computer Science and Software Engineering, Shenzhen University. He was a Postdoctoral Fellow with Institute of Network Coding (INC) of The Chinese University of Hong Kong (CUHK) from Aug. 2012 to Aug. 2013. He won his PhD degree and Mphil degree from Information Engineering (IE) Department of CUHK respectively. He got Bachelor degree in Computer Science from University of Science and Technology of China (USTC). His research interests lie in distributed/federated learning, privacy protection and caching algorithm design in networks. He has published more than 90 papers including IEEE INFOCOM, ICNP, IWQoS, IEEE ToN, JSAC, TPDS, TMC, TMM, etc.
\end{IEEEbiography}

\end{document}